\documentclass[runningheads]{llncs}

\usepackage{cite}
\usepackage{amsmath,amssymb,amsfonts}
\usepackage{algorithm}
\usepackage[noend]{algorithmic}
\usepackage{graphicx}
\usepackage{subfig}
\usepackage{multirow}
\usepackage{threeparttable}
\usepackage{tabularx}
\usepackage{booktabs}
\usepackage{url}
\usepackage{textcomp}
\usepackage{wrapfig}
\usepackage{xcolor}
\usepackage[colorlinks,
            linkcolor={black},
            citecolor={blue},
            urlcolor={black},
            pdfauthor={No Author Given},
            bookmarks=false,
            pdfsubject={Random Similarity Forests},
            pdfkeywords={complex data, random forests, distance-based classifier, multi-omics}
            ]{hyperref}
            
\newcommand{\textapprox}{\raisebox{0.5ex}{\texttildelow}}
\newcolumntype{R}{>{\raggedleft\arraybackslash}X}%
\DeclareMathOperator*{\argmin}{\arg\!\min}

\begin{document}

\title{Random Similarity Forests}

\author{Maciej Piernik\inst{1,2}\orcidID{0000-0001-7281-284X} \and
Dariusz Brzezinski\inst{1,2,3}\orcidID{0000-0001-9723-525X} \and
Pawel Zawadzki\inst{2,4}\orcidID{0000-0002-9032-2315}}

\authorrunning{M. Piernik et al.}

\institute{
Institute of Computing Science, Faculty of Computing and Telecommunications, Poznan University of Technology,
Piotrowo 2, 60-965 Poznan, Poland \and
MNM Bioscience Inc., 1 Broadway, Cambridge, MA 02142 \and
Institute of Bioorganic Chemistry of the Polish Academy of Sciences, Zygmunta Noskowskiego 12/14, 61-704 Poznan, Poland \and
Physics/Biology Department, Faculty of Physics, Adam Mickiewicz University, Uniwersytetu Poznanskiego 2, 61-614 Poznan, Poland
\email{\{maciej.piernik,dariusz.brzezinski,pawel.zawadzki\}@mnm.bio}}

\maketitle

\begin{abstract}
The wealth of data being gathered about humans and their surroundings drives new machine learning applications in various fields. Consequently, more and more often, classifiers are trained using not only numerical data but also complex data objects. For example, multi-omics analyses attempt to combine numerical descriptions with distributions, time series data, discrete sequences, and graphs. Such integration of data from different domains requires either omitting some of the data, creating separate models for different formats, or simplifying some of the data to adhere to a shared scale and format, all of which can hinder predictive performance. In this paper, we propose a classification method capable of handling datasets with features of arbitrary data types while retaining each feature's characteristic. The proposed algorithm, called Random Similarity Forest, uses multiple domain-specific distance measures to combine the predictive performance of Random Forests with the flexibility of Similarity Forests. We show that Random Similarity Forests are on par with Random Forests on numerical data and outperform them on datasets from complex or mixed data domains. Our results highlight the applicability of Random Similarity Forests to noisy, multi-source datasets that are becoming ubiquitous in high-impact life science projects.
\end{abstract}

\section{Introduction}
\label{sec:introduction}
Over the last decades, machine learning has matured to the point in which researchers are spoiled for choice with classifiers for \textit{scalar} data.
There are hundreds of classifiers to choose from and their performance has been tested on thousands of datasets spanning across various domains~\cite{Fernandez:2014}.
Increasingly, however, researchers are more and more often faced with datasets of \textit{complex} composition, where numerical features are not explicitly available.
In such cases, there are two main options to choose from.
Either to transform complex objects into many simple features, which can then be used by any traditional feature-based classifier (e.g., Random Forests~\cite{Breiman:2001}) or a dedicated one (e.g., CNNs for images~\cite{LeCun:2015}), or to rely on a domain-specific distance measure and use distance-based classifiers (e.g., Similarity Forests~\cite{Sathe:2017}).
However, there is a third option that relies on the fact that many complex data types can be broken down into both traditional features (e.g., numerical or categorical) and simpler but still complex substructures for which good distance measures exist.
For example, sequences can be decomposed into subsequences, sets of elements, distributions of elements, or distributions of lengths of subsequences.
These complex substructures can then be combined together with simple, numerical or categorical, features and form a \textit{mixed} type dataset to better capture the detailed characteristics of each example.
Unfortunately, the currently available classification methods inherently enable only one of the first two types of analyses, on scalar data or complex objects.
This limitation may prohibit us from using the full potential hidden in the data, as it forces us to either reduce all complex structures into simple features (most probably with information loss) or process all features with a single, complex distance measure (which is prone to the curse of dimensionality).

Consider a scenario of building a tumor genome classifier predicting patients responding to a given treatment based on whole-genome sequencing (WGS)~\cite{Davies:2017}.
Although, theoretically, a distance-based approach can be used in this scenario, it is unlikely to produce good results because of the curse of dimensionality and computational infeasibility (as the reference human genome is \textapprox3.2 billion base pairs long).
Moreover, this approach also neglects decades of research in the field of oncology and genomics, which provide a wide array of potentially useful biomarkers.
A more natural approach in this scenario is to use a feature-based classifier that can make use of existing biomarkers as well as explore new ones.
Currently however, this approach necessitates a drastic decomposition of the data into a fixed set of scalar features.
At the same time, the information in the human genome can be decomposed into many intermediate structures carrying significantly more information than their scalar counterparts.
Examples of such structures include variant distributions, sequential patterns, gene sets, interaction graphs.
All of these structures have multiple distance measures available and could be used in conjunction with existing numerical biomarkers as a single dataset to produce more informed predictions.
However, to the best of our knowledge, there does not exist any classifier capable of handling data with mixed, arbitrarily defined types of features.

In this paper, we propose a new classifier, called Random Similarity Forests (RSF), capable of handling datasets with mixed, arbitrarily defined feature types.
Our approach requires a distance measure to be provided for each feature and works as a blend of Random Forests and Similarity Forests.
By using a mixture of feature types, Random Similarity Forests inherit the benefits of both feature-based and similarity-based methods. 
In the paper, we intuitively describe and formally define the proposed classifier. We also perform sensitivity tests with respect to the number of features analyzed in each tree node and the number of trees in the forest. The method is then experimentally evaluated against Random Forests and Similarity Forests on datasets with scalar features, complex objects with a distance measure, and mixed, arbitrarily defined features with distance measures. Our analysis highlights the characteristic properties of each approach and discusses their suitability for different dataset types.

\section{Related work}
\label{sec:related}
Random forests~\cite{Breiman:2001} are one of the most popular classifiers and have been shown to offer the best quality of predictions across a wide range of datasets when compared to other classifiers~\cite{Fernandez:2014}.
However, like many other approaches, their use is limited to scalar data and, therefore, requires feature extraction when dealing with complex objects.
Due to this fact, Similarity Forests~\cite{Sathe:2017} have been recently proposed as an alternative to Random Forests, especially when regular features are absent but similarities or distances between examples are attainable.

When dealing with distance-based classification, kernel methods are a popular group to consider, especially since there have been studies exploring the relationship between Random Forests and kernel methods~\cite{Scornet:2016}.
In general, distance-based classification is often applied when dealing with complex data, e.g., for microbiome, image, or time series classification. A different way of dealing with complex data in classification using distances is through a combination of clustering and distance-based feature extraction.
This approach relies on unsupervised clustering of the data and later encoding the clusters as features, either through aggregate inter- and intra-cluster distances~\cite{Tsai:2011} or simply by encoding the distances to each cluster center as separate features~\cite{Piernik:2021}.

Recently, a different combination of feature-based and distance-based methods has been proposed by Liang \textit{et al.}~\cite{Liang:2019}, where the authors first use Random Forests to select the most important features according to their impurity-based importance and later use these features in a k-Nearest Neighbors classifier with dynamically selected distance measures.

Although the described existing approaches create forests or calculate distances between objects, they either require scalar features or are based on a single distance calculated for entire examples. Similarly, integrative approaches from biomedical studies that combine data from different domains either require a common feature representation, create separate data type specific models, or perform data re-scaling and simplification~\cite{multiomics}.
To the best of our knowledge, Random Similarity Forests are the only single-classifier model capable of dealing with datasets with a mixture of arbitrarily defined features without any data transformations.

\section{Random Similarity Forest}
\label{sec:main}

\subsection{Preliminaries}
\label{sec:main:preliminaries}
Given a dataset of \textit{training examples} $\mathcal{X}=\{\pmb{x_1}, \pmb{x_2}, \ldots, \pmb{x_n}\}$ and their corresponding \textit{class labels} $\pmb{y}=\{y_1, y_2, \ldots, y_n\}$, the task of a \textit{classifier} is to predict the class label $\hat{y}$ of each \textit{unlabeled example} $\pmb{\hat{x}}$ (i.e., an example for which the class is unknown).
Every example $\pmb{x_i}$ is a list $(x_{i1}, \ldots, x_{ij}, \ldots ,x_{ip})$ of the same length $p$, where each position $j$ holds a value a given example has for a particular \textit{feature} $F_1, F_2, \ldots, F_p\in \mathcal{F}$.
We also denote all values of $\mathcal{X}$ across a given feature $F_j$ by $\pmb{x_{.j}}$. Each class label $y_i$  falls into one of several categorical class values $\mathcal{C} = \{c_1, c_2, \ldots, c_m\}$.
This is the standard definition of a classification problem.
In our case, each feature can be from a different, arbitrary data domain.
Additionally, every feature $F_1, F_2, \ldots, F_p\in \mathcal{F}$ has an associated distance measure $\delta_1, \delta_2, \ldots , \delta_p\in\Delta$. Therefore, the distance between any two examples $\pmb{x_i}$, $\pmb{x_j}$ for a given feature $F_l$ is equal to $\delta_l(x_{il}, x_{jl})$, or $\delta_l(\pmb{x_i}, \pmb{x_j})$ for short.

\subsection{The Algorithm}
\label{sec:main:algorithm}
Just like Random Forests and Similarity Forests, Random Similarity Forests rely on bagging of a user-defined number (\textit{max\_trees}) of single-tree classifiers.
A single Random Similarity Tree is constructed in a top-down fashion by recursively splitting the examples in each node into two exclusive subsets.
The recursion stops when a given node contains only examples from a single class or when one of the early-stopping conditions is met (discussed in Section~\ref{sec:main:discussion}).
Each split is calculated using a similarity-based 1-dimensional projection of the examples in a given node.
First, a subset of \textit{max\_features} candidate features is selected, and afterward, \textit{max\_pairs} pairs of examples are picked at random for each feature.
The examples in a pair are selected so that they come from different classes and have a different value on a given feature.

For a given pair of examples $\pmb{x_p}$, $\pmb{x_q}$, the distance between all other examples $\pmb{x_i}\in\mathcal{X}_{\mathit{node}} \setminus \{\pmb{x_p}$, $\pmb{x_q}\}$ and the two selected examples is evaluated for each selected feature $F_l$ in order to create a projection $\mathcal{P}(\pmb{x_i})$ of each example $\pmb{x_i}$ into a direction defined by $\pmb{x_p}$ and $\pmb{x_q}$.
As shown by Sathe and Aggarwal~\cite{Sathe:2017}, for the purposes of constructing a split, this projection can be approximated with $\mathcal{P}(\pmb{x_i})\propto \delta_l(\pmb{x_q}, \pmb{x_i})-\delta_l(\pmb{x_p}, \pmb{x_i})$, as this approximation preserves the order of the original projection and the order is the only information necessary to construct a split.
Therefore, for each pair of examples $\pmb{x_p}$, $\pmb{x_q}$, all remaining examples $\pmb{x_i}$ in a given node are ordered by $\delta_l(\pmb{x_q}, \pmb{x_i})-\delta_l(\pmb{x_p}, \pmb{x_i})$ along a given feature $F_l$.

This projection approximation can be viewed as a dynamic feature. After the projection has been computed, it can be used to perform a split just like any typical numerical feature would be used in a regular decision tree.
The split point is selected as the point which minimizes the weighted average of the Gini index of the children nodes.
For a given node $N_i$, the Gini index is defined as $G(N_i)=1-\sum_{c\in\mathcal{C}}{p_c^2}$, where $p_c$ is the fraction of examples of class $c$ in a given node.
Therefore, given two nodes $N_i$, $N_j$ consisting of $n_i$, $n_j$ examples, respectively, the weighted Gini index is calculated as:
\begin{equation}
    G(N_i, N_j)=\frac{n_iG(N_i) + n_jG(N_j)}{n_i+n_j}\text{.}
    \label{eq:gq}
\end{equation}
Here, we opted for the Gini-index-based splitting strategy used also in Similarity Forests, however, it is important to note that after the projection is done we are in fact in possession of a dynamically calculated numerical feature. Therefore, other measures, such as entropy, can also be used.

Given the order defined with the projection and an impurity measure, all possible splits along a given feature are evaluated and the split that minimizes the impurity is selected.
After performing this procedure for all selected features in a given node, the feature producing the best split is selected.
Subsequently, two new tree nodes are created and the examples falling below the split threshold are placed in one node while the remaining examples are placed in the other node, and the whole procedure is repeated until all leaf nodes are pure or an early-stopping criterion is met.

\begin{figure}[!htb]
	\centering
		\includegraphics[width=3.4in]{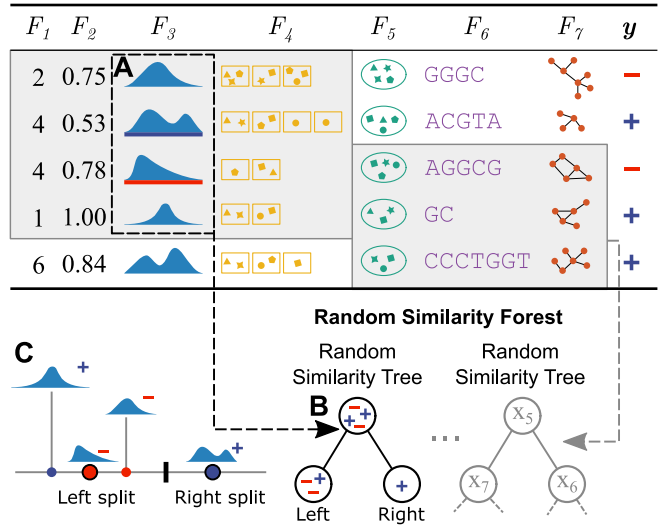}
	\caption{An example illustrating the training process of Random Similarity Forests. Gray areas symbolize the bootstrap samples and subset of features going into each tree. Feature $F_3$ (\textbf{\textsf{A}}) is selected to perform a split in one of the trees (\textbf{\textsf{B}}). The distance-based projection (\textbf{\textsf{C}}) for that feature is calculated and the split point yielding the purest split is selected. The reference points (distributions) $\pmb{x_p}, \pmb{x_q}$ used to create the projection are underlined in red and blue in the table.}
	\label{fig:example:rsf}
\end{figure}

The above-described process is illustrated in Fig.~\ref{fig:example:rsf}. The pseudocode of training a single Random Similarity Tree is presented in Algorithm~\ref{alg:rsf}.

\begin{algorithm}[ht]
    \scriptsize
    \caption{Random Similarity Tree: $\mathit{RST}(\cdot)$}
    \label{alg:rsf}
\begin{algorithmic}[1]
    \REQUIRE $\mathcal{X}$: training examples; $\pmb{y}$: examples' class labels; $\pmb{\delta}$: distance measures for each feature; \textit{max\_features}: number of features picked randomly at each node; \textit{max\_pairs}: number of example pairs picked at random for each feature at each node
    \ENSURE a trained Random Similarity Tree model
    \IF{$\mathit{earlyStopping()}$ \OR $len(unique(\pmb{y})) = 1$}\label{l:stop}
        \STATE{$\mathit{setLeaf()}$}
        \RETURN{$\mathit{self}$}
    \ELSE
        \FOR{$1 .. \mathit{max\_features}$}\label{l:features}
            \STATE{$F_j\leftarrow \mathit{random}(\{F_j\in\mathcal{F}:\mathit{var}(\pmb{x_{.j}}) > 0\})$}\label{l:f_pick}
            \STATE{$c_1\leftarrow \argmin_{c\in \mathit{unique}(\pmb{y})}\{\mathit{var}(\{x_{ij}: y_i = c\})\}$}\label{l:c1}
            \STATE
            \FOR{$1 .. \mathit{max\_pairs}$}\label{l:pairs}
                \STATE{$\pmb{x_p} \leftarrow \mathit{random}(\{\pmb{x_i}: y_i = c_1\})$}
                \STATE{$\pmb{x_q} \leftarrow \mathit{random}(\{\pmb{x_i}: y_i \neq c_1 \wedge x_{ij} \neq x_{pj}\})$}\label{l:p_pick}
                \FOR{$\pmb{x_i} \in \mathcal{X}$}\label{l:proj}
                    \STATE{$P(\pmb{x_i})\leftarrow\delta_j(\pmb{x_q}, \pmb{x_i}) - \delta_j(\pmb{x_p}, \pmb{x_i})$}
                \ENDFOR\label{l:proj:end}
                \FOR{$\mathit{thr}\in \mathit{unique}(P)$}\label{l:find}
                    \STATE{$\pmb{y_{\mathit{left}}}\leftarrow \{y_i\in\pmb{y}: P(\pmb{x_i}) \leq \mathit{thr}\} $}\label{l:split:l}
                    \STATE{$\pmb{y_{\mathit{right}}}\leftarrow \{y_i\in\pmb{y}: P(\pmb{x_i}) > \mathit{thr}\} $}\label{l:split:r}
                    
                    \STATE{$g\leftarrow G(\pmb{y_{\mathit{left}}}, \pmb{y_{\mathit{right}}})$}\label{l:gini}
                    \IF{$(g < best_g)$ \OR \\\quad($g = best_g$ \AND \\\quad\quad $|len(y_{\mathit{left}})-len(y_{\mathit{right}})|<best_{bal}$)}\label{l:store}
                        \STATE{$update\_best(g, \mathit{thr}, F, \pmb{x_p}, \pmb{x_q}, P, bal)$}
                    \ENDIF\label{l:store:end}
                \ENDFOR\label{l:find:end}
            \ENDFOR
        \ENDFOR
        \STATE
        \STATE{$\mathcal{X}_{\mathit{left}}, \pmb{y_{\mathit{left}}}\leftarrow \{\pmb{x_i}\in\mathcal{X}, y_i\in\pmb{y}: P(\pmb{x_i}) \leq \mathit{thr}\} $}\label{l:split:fl}
        \STATE{$\mathcal{X}_{\mathit{right}}, \pmb{y_{\mathit{right}}}\leftarrow \{\pmb{x_i}\in\mathcal{X}, y_i\in\pmb{y}: P(\pmb{x_i}) > \mathit{thr}\} $}\label{l:split:fr}
        \STATE{$N_{\mathit{left}}\leftarrow \mathit{RST}(\mathcal{X}_{\mathit{left}}, \pmb{y_{\mathit{left}}}, \pmb{\delta}, \mathit{max\_features}, \mathit{max\_pairs})$}\label{l:node:l}
        \STATE{$N_{\mathit{right}}\leftarrow \mathit{RST}(\mathcal{X}_{\mathit{right}}, \pmb{y_{\mathit{right}}}, \pmb{\delta}, \mathit{max\_features}, \mathit{max\_pairs})$}\label{l:node:r}
    \ENDIF

    \RETURN{$\mathit{self}$}
\end{algorithmic}
\end{algorithm}

After all \textit{max\_trees} Random Similarity Trees are computed, the trained Random Similarity Forest is ready to make predictions.
For each node in each tree, given an unlabeled example $\pmb{\hat{x}}$, the example is projected along the feature and the two examples stored in that node and assigned to one of the child nodes based on the stored split point.
Following this procedure, once the example reaches a leaf node, it returns a weighted prediction of the majority class in that node.
After predictions from all trees in the forest are made, the example is assigned to the class with the highest weighted average from all single-tree predictions.

\subsection{Computational Complexity}
\label{sec:main:complexity}
Let us now discuss the computational complexity of Random Similarity Forests (RSF).
Unsurprisingly, it is strongly related to the complexity of Similarity Forests (SF), which is $\mathcal{O}(n\cdot\log{n})$.
However, RSF additionally adds the cost of checking multiple features at each node, so the complexity is raised by that factor to $\mathcal{O}(p\cdot n\cdot\log{n})$, which is equivalent to the complexity of Random Forests (RF).
However, it is worth noting that in practice the value of the \textit{max\_features} parameter for RSF may need to be higher than for RF as it is inherently more random as it randomly picks both features and pairs of examples.
Moreover, even though the number of pairs is a parameter, we recommend picking only a single pair for each feature, as suggested by Sathe and Aggrawal~\cite{Sathe:2017}.

Another cost hidden in the computational complexity is the cost of calculating the distances between examples.
It is an essential factor because, depending on the feature type, it can either be negligible or a very costly operation.
For simple numerical features, the distance computation can be usually omitted as for any metric it would produce a projection equivalent to the original feature in terms of the example order.
On the other hand, more complex data types may require substantial computation, e.g., edit distance has quadratic complexity for sequential data and is NP-complete for graphs.
Therefore, the use of such metrics may be prohibitive for longer sequences or larger graphs.

\subsection{Discussion}
\label{sec:main:discussion}
Since Random Similarity Forests draw from Random Forests and Similarity Forests, they combine their advantages and possess unique properties:

\begin{enumerate}
    \item By using feature-oriented similarity metrics to split data points, Random Similarity Forests can be used to classify datasets characterized by numerical as well as complex data features. Random Forests cannot classify complex data objects, whereas Similarity Forests treat entire examples as data objects, hiding the  characteristics of individual features.
    \item By sampling feature subsets for each node split and by analyzing each feature separately, Random Similarity Forests are robust to noisy datasets with a lot of features. Similarity Forests treat entire examples as data objects, which can hinder their performance on datasets with many irrelevant features.
    \item Analyzing similarities based on each feature separately allows Random Similarity Forests to use several similarity measures to analyze the same objects and let the forest decide which measures are the most useful for classification.
\end{enumerate}

In addition to combining the advantages from both Random Forests and Similarity Forests, Random Similarity Forests also inherit some of their limitations.
The first limitation comes from Similarity Forests and it is the fact that it is designed for binary classification problems.
However, one can easily adapt it to multi-class classification or regression problems analogously to the method proposed for Similarity Forests~\cite{Czekalski:2021}, or by using the one-vs-all approach.

Another characteristic worth discussing is the effect of feature variance on node splits. If one would select the pair of examples $(\pmb{x_p}$, $\pmb{x_q})$ with the same value of feature $F_j$, then they  would  produce  a completely  random  ordering of examples for the split (since if $x_{pj} = x_{qj}$ then $\forall{\pmb{x_i}}:  \delta_j(\pmb{x_q}, \pmb{x_i}) - \delta_j(\pmb{x_p}, \pmb{x_i}) = 0$). Similar problems appear in Random Forests when the feature values are the same for examples of different classes, whereas Similarity Forests do not suffer from this particular limitation so often as they always consider each example as a whole. For datasets with many features with low variance, this reduces the effective number of features tested at each node and promotes high-variance features. That is why Random Similarity Forests will perform better on datasets with features with high variance. This limitation can be overcome by tuning the \textit{max\_features} hyperparameter. As shown by Geurts \textit{et al.}~\cite{Geurts:2006}, the best value for this parameter highly depends on the number of correlated and irrelevant features: lower values work better with highly correlated features while higher values  with irrelevant features. Indeed, as will be shown in Section~\ref{sec:experiments:sensitivity}, the proposed default value of \textit{max\_features} for Random Similarity Forests is $0.5p$, which is higher than $\sqrt{p}$ commonly used for Random Forests~\cite{Geurts:2006}.

Other issues worth mentioning are the early-stopping conditions (pre-pruning) and pruning (post-pruning). There are many possible options to stop the tree construction process before it produces perfectly pure leaf nodes. These could include, e.g., limiting the maximum depth a tree can reach, specifying the minimum number of elements required to perform a split or the minimum number of elements to form a leaf node. Similarly, there are multiple ways in which a tree can be pruned, with arguably the most popular method being minimal cost complexity pruning~\cite{cart}. However, pre-and post-pruning methods are not specific to the proposed algorithm or any other forest algorithm and are out of the main scope of this paper.

\section{Experimental evaluation}
\label{sec:experiments}
In this section, we experimentally evaluate the Random Similarity Forests algorithm and illustrate its usefulness in real-world scenarios.
Since the proposed classifier is related to both Random Forests and Similarity Forests, the nature of the evaluation is highly comparative. Our goal is to showcase the strengths and limitations of our proposal and indicate when its use would be the most beneficial and when it might be preferable to rely on one of the alternatives.
First, we compare the classification performance of the three classifiers on ten publicly available datasets with simple numerical features.
Next, we compare the three approaches on complex data, i.e., datasets consisting of arbitrarily defined objects with distance measures. For this purpose, we use sequences of sets, time series, and graph datasets.
Finally, we illustrate the usefulness of our proposal by training the classifiers on datasets consisting of mixtures of complex object-like features and simple numerical features.

\subsection{Experimental Setup}
In all of the experiments, we compare Random Similarity Forests (RSF) against Random Forests (RF) and Similarity Forests (SF) using the area under the ROC curve (AUC). We chose AUC since we focus on binary classification, it is skew invariant, assesses the classifiers' ranking abilities, and it has been shown to be statistically consistent and more discriminating than accuracy~\cite{Huang:2005}.
To evaluate each approach on each dataset, we rely on 10 repetitions of stratified 2-fold cross-validation (10$\times$2CV).
Afterwards, the results undergo a series of statistical tests~\cite{Demsar:2006}.
We rely on the Friedman statistic with a post-hoc Nemenyi test to distinguish between the compared approaches across all datasets.
Afterwards, we check for differences on each individual dataset.
First, we verify the normality assumption using the Shapiro-Wilk test.
If the assumption is met, we use the ANOVA statistic with a post-hoc Tukey HSD test.
If the assumption is not met, we once again resort to the non-parametric Friedman statistic with a post-hoc Nemenyi test~\cite{Demsar:2006}. All tests were performed at significance level $\alpha = 0.05$.

The code used to perform the experiments was written in Python with parts of the implementation of Random Similarity Forests written in Cython to achieve higher efficiency.
For Random Forests, we use the scikit-learn implementation~\cite{scikit-learn}, whereas for Similarity Forests, we rely on the implementation described in~\cite{Czekalski:2021}.
Unless stated otherwise, we rely on the default hyperparameters for each classifier, as specified in their implementations.

\subsection{Datasets}
Since our experimental analysis focuses on data with scalar, complex, and mixed types of features, we use three groups of datasets, each corresponding to a given type of features.
Additionally, we have a separate set of datasets dedicated to performing sensitivity analyses. The choice of datasets was made a priori and independently of the results obtained with our methods.

\subsubsection*{Datasets for sensitivity analysis}
For sensitivity analysis, we used binarized versions of 12 numerical datasets: ten from the UCI Machine Learning\footnote{\url{https://archive.ics.uci.edu/ml/}} repository and two introduced by Breiman; for a detailed description of these datasets see~\cite{Geurts:2006}. We chose these datasets as they were used to discuss the impact of different hyperparameter values for Random Forests and Extra Trees~\cite{Geurts:2006}. Importantly, these datasets are independent of those used for assessing the predictive performance of RF, SF, and RSF, to avoid biasing the comparative analysis.

\subsubsection*{Scalar data}
For the analysis of scalar data, we used ten publicly available datasets with numeric features.
The datasets exhibit various conditions in terms of the number of features, learning sample size, and feature variance, and can be accessed through the LIBSVM dataset repository.\footnote{\url{https://www.csie.ntu.edu.tw/~cjlin/libsvmtools/datasets/}}
We used the scaled versions of the datasets, as required by Similarity Forests.
Each dataset was additionally preprocessed by removing features containing missing values.
Table~\ref{tab:datasets} presents the main characteristics of each dataset after preprocessing.

\subsubsection*{Complex data}
For the analysis of complex data (i.e., data where each example is an object without any explicitly defined numerical features), we used three groups of datasets consisting of sequences of sets, time series, and graphs, respectively.

Sequences of sets are a good candidate for complex data because they combine set information, which can be easily encoded with numerical features, and sequential information, which is very difficult to encode as scalars but manifests itself through distance measures on whole objects. Therefore, we used a synthetic data generator\footnote{\url{https://gingerbread.shinyapps.io/SequencesOfSetsGenerator/}} to create three  datasets of sequences of sets: \texttt{items}, \texttt{lengths}, and \texttt{order}. The sequences in the \texttt{items} dataset have the same mean length and mean set size, and are differentiated by the distributions of the selected items between the two classes. Sequences in the \texttt{lengths} dataset are generated with identical set size and item distributions but are differentiated by the lengths between the classes. Finally, all the sequences in the \texttt{order} dataset are generated from the same length, set size, and item distributions, and are only differentiated by the order of the elements in one of the classes.

Time series datasets were taken from the UCR Time Series Classification Archive\footnote{\url{https://www.cs.ucr.edu/~eamonn/time_series_data_2018/}}, with distances between objects calculated with dynamic time warping. The graph datasets were taken from the TU-Dortmund Graph Kernel Benchmarks\footnote{\url{https://ls11-www.cs.tu-dortmund.de/staff/morris/graphkerneldatasets}} with distances between objects calculated using the Ipsen-Mikhailov distance. In both cases, we limited our selection to binary classification problems, with varying numbers of examples and features (length, sets, nodes, edges).

\subsubsection*{Mixed data}
For the analysis of mixed data (i.e., data where each example is described by arbitrarily defined features with a distance measure for each feature), we used four groups of datasets consisting of time series, graphs, multi-omics data, and genome sequencing data.

Time series datasets consisted of three distance measures (euclidean, cosine, dynamic time warping) computed for series from the UCR Repository.\footnote{\url{https://www.cs.ucr.edu/~eamonn/time_series_data_2018/}} Similarly, the graph datasets consisted of four distance measures (portrait divergence, Jaccard distance, degree divergence, Ipsen-Mikhailov distance) computed for graphs from the TU-Dortmund repository\footnote{\url{https://ls11-www.cs.tu-dortmund.de/staff/morris/graphkerneldatasets}} using the netrd package~\cite{netrd}. The multiomics datasets consisted of Alzheimer's Disease and BRCA-mutated breast cancer patients, described by three sets of numeric expression data.\footnote{\url{https://github.com/txWang/MOGONET}} Finally, the last group of datasets represented tumor samples of ovarian and breast cancer patients obtained using whole genome sequencing.\footnote{\url{https://github.com/MNMdiagnostics/dbfe}} In this group of datasets, each sample is characterized by 50 gene amplifications represented as simple numerical features and 6 distributions of lengths of various DNA structural variants.

\subsection{Sensitivity Analysis}
\label{sec:experiments:sensitivity}

\begin{figure*}[!htb]
	\centering
		\includegraphics[width=\textwidth]{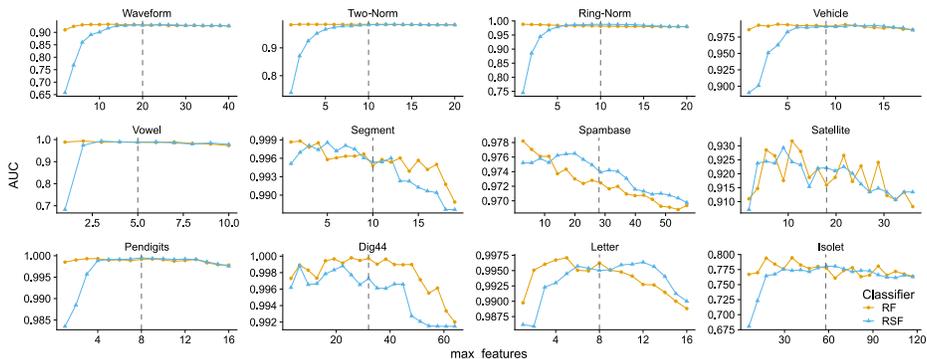}
	\caption{Evolution of mean AUC of Random Forests (RF) and Random Similarity Forests (RSF) with varying \textit{max\_features} on 12 datasets. Dashed line shows the proposed RSF default value of \textit{max\_features}=$0.5p$.}
	\label{fig:results:sensitivity}
\end{figure*}

Using the 12 datasets proposed in~\cite{Geurts:2006}, we have analyzed the effect of \textit{max\_features} on Random Similarity Forests. The parameter denotes the number of potential splits screened at each node during the growth of a Random Similarity Tree. It may be chosen in the interval $[1,\ldots,p]$ and the smaller its value the stronger the randomization of the trees. Figure~\ref{fig:results:sensitivity} compares the AUC of Random Forests (RF) and Random Similarity Forests (RSF) for increasing values of \textit{max\_features}.

One can notice that for all the datasets the performance of RSF first increases and then declines as \textit{max\_features} approaches the total number of features in a dataset $p$. As mentioned in Section~\ref{sec:main:discussion}, this phenomenon is related to the number of correlated and irrelevant features in a dataset: lower values work better with highly correlated features while higher values of \textit{max\_features} work better with irrelevant features. This is due to the fact that a higher value of \textit{max\_features} leads to a better chance of filtering out the irrelevant variables. These findings are in accordance with~\cite{Geurts:2006}, were similar trends were noticed for the Extra Trees classifier. It is worth noting, however, that compared to RF, the increase of predictive performance with  \textit{max\_features} is slower. This stems from the fact that RSF samples pairs of examples and creates different projections, thus introducing additional randomness. Therefore, we propose to use a default value of  \textit{max\_features}=$0.5p$, as opposed to $\sqrt{p}$, which is the commonly used default for Random Forests and Extra Trees~\cite{Geurts:2006}.

\begin{wrapfigure}{r}{0.52\textwidth}
	\centering
		\includegraphics[width=0.50\textwidth]{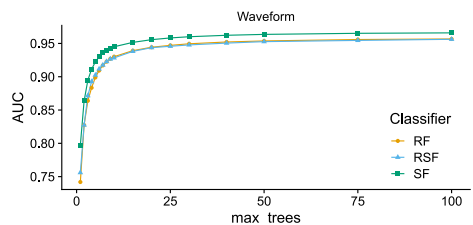}
	\caption{Mean AUC of Random Forests (RF), Random Similarity Forests (RSF) and Similarity Forests (SF) with varying \textit{max\_trees} on the Waveform dataset.}
	\label{fig:results:sensitivity_m}
\end{wrapfigure}

We have also analyzed the \textit{max\_trees} parameter, responsible for the number of trees in the ensemble (Fig. \ref{fig:results:sensitivity_m}).
The results are in accordance with studies on different tree ensembles~\cite{Geurts:2006}: the higher the value of \textit{max\_trees}, the better the ensemble's accuracy.
It can be noticed that all three algorithms (RF, SF, RSF) have the same speed of convergence.
In the remaining experiments, for all three classifiers, we use \textit{max\_trees}=100, which is large enough to ensure convergence of the ensemble effect on all our datasets.

\subsection{Evaluation on Scalar Data}
The aim of the first set of experiments was to assess the predictive performance of Random Similarity Forests (RSF) on scalar data. Since scalar features are the natural data domain of Random Forests (RF), the aim is also to verify whether the proposed classifier will be able to match the performance of Random Forests.
Since Similarity Forests (SF) are also capable of dealing with scalar data and were reported to produce results on par with or even better than Random Forests~\cite{Sathe:2017}, we include this approach in this comparison as well. The results of these experiments are presented in Table~\ref{tab:datasets}.

\begin{table}[ht!]
\scriptsize
\begin{threeparttable}[b]
\centering
\caption{Dataset characteristics and mean AUC performance of Random Forests (RF), Similarity Forests (SF), and Random Similarity Forests (RSF) from 10 $\times$ 2-fold cross-validation experiments. Features for sequences of sets denoted as [mean set size]/[mean length]. Features for graphs denoted as [mean number of nodes]/[mean number of edges].}
\begin{tabularx}{\textwidth}{l@{\quad}c@{ }crrcXXX}
\toprule
\multirow{2}{*}{Dataset} & \multirow{2}{*}{Type} & \multirow{2}{*}{Features} & \multirow{2}{*}{\#Ex.} & \multirow{2}{*}{\#Features} & \multirow{2}{*}{Distance} & \multicolumn{3}{c}{AUC}\\
\cmidrule{7-9}
& & & & & & RF & RSF & SF \\
\midrule
\texttt{australian} & \multirow{10}{*}{scalar}   & \multirow{10}{*}{numeric}   & 690  & 8  &  \multirow{10}{*}{euclidean}  & \textbf{0.92}$^\circ$ &	0.91 &	0.92 \\
\texttt{breast}       &  &                          & 683  & 10 &   & 0.99 &	0.99 &	\textbf{1.00}$^\circ$ \\
\texttt{diabetes}     &  &                          & 768  & 6  &   & \textbf{0.80} &	0.80 &	0.80 \\
\texttt{german}       &  &                          & 1000 & 18 &   & \textbf{0.73} &	0.73 &	0.73 \\
\texttt{heart}        &  &                          & 270  & 9  &   & 0.88 &	0.87 &	\textbf{0.89}$^\circ$ \\
\texttt{ionosphere}   &  &                          & 351  & 2  &   & 0.84 &	\textbf{0.84} &	0.84 \\
\texttt{liver}        &  &                          & 145  & 4  &   & \textbf{0.77} &	0.75 &	0.76 \\
\texttt{sonar}        &  &                          & 208  & 59 &   & 0.89 &	0.88 &	\textbf{0.91}$^\circ$ \\
\texttt{splice}       &  &                          & 1000 & 60 &   & 0.99$^\circ$ &	\textbf{0.99}$^\circ$ &	0.90 \\
\texttt{svmguide3}    &  &                          & 1243 & 21 &   & 0.84$^\circ$ &	\textbf{0.85}$^\circ$ &	0.77 \\
\midrule
\texttt{items}  &  \multirow{10}{*}{complex}     & \multirow{3}{*}{\parbox{2cm}{\centering sequences of sets}} & 400  & 20/20 &  \multirow{3}{*}{\parbox{2cm}{\centering edit\\distance}} & \textbf{1.00}$^{\bullet b}$ & 0.91$^\circ$ &	0.85 \\
\texttt{length}      &  &                          & 400  & 20/20 &   & 0.86$^b$ & 0.91$^\circ$ &	\textbf{0.92}$^\circ$ \\
\texttt{order}       &  &                          & 400  & 20/20 &   & 0.56$^b$ & \textbf{0.79}$^\bullet$ &	0.73$^\circ$ \\
\cmidrule{3-9}
\texttt{computers}   &  &  \multirow{3}{*}{\parbox{2cm}{\centering time series}}                      & 500  & 720 &  \multirow{3}{*}{\parbox{2cm}{\centering dynamic time warping}} &  0.68  & 0.77$^\circ$ &	\textbf{0.78}$^\circ$ \\
\texttt{housetwenty}   &  &                        & 159  & 2000 &   &  0.92  & 0.99$^\circ$ &	\textbf{0.99}$^\circ$ \\
\texttt{toeseg} & &                        & 268  & 277 &   &  0.85  & 0.96$^\circ$ &	\textbf{0.97}$^\circ$ \\
\cmidrule{3-9}
\texttt{cox2}           & &  \multirow{4}{*}{\parbox{2cm}{\centering graph}}                      & 467  & 41/43 &  \multirow{4}{*}{\parbox{2cm}{\centering ipsen-mikhailov}}       &  0.65$^b$   & 0.70$^\circ$ &	\textbf{0.71}$^\circ$ \\
\texttt{mutag}          & &                        & 188  & 18/20 &   &  0.87$^b$   & \textbf{0.92}$^\circ$ &	0.92$^\circ$ \\
\texttt{proteins}       & &                            & 1113  & 39/73  &   &  0.77$^b$   & \textbf{0.95}$^\bullet$ &	0.91$^\circ$ \\
\texttt{ptcfm}          & &                            & 349  & 14/14  &   &  0.58$^b$   & \textbf{0.63}$^\circ$ &	0.61$^\circ$ \\

\midrule
\texttt{beetlefly} & \multirow{10}{*}{mixed}  & \multirow{2}{*}{time series} & 40  & 512 & \multirow{2}{*}{\parbox{2cm}{\centering euclidean, cosine, dtw}} & 0.84 & \textbf{0.85} & 0.81$^{m}$ \\
\texttt{birdchicken} &   & & 40  & 512 &   & 0.89 & \textbf{0.90} & 0.87$^{m}$  \\
\cmidrule{3-9}
\texttt{bzr}         & & \multirow{2}{*}{graphs}   & 405  & 36/38 & \multirow{2}{*}{\parbox{2cm}{\centering portrait, degree, jaccard, ipsen}}  & 0.73$^b$ & \textbf{0.90}$^\bullet$ &	0.87$^{m\circ}$ \\
\texttt{dhfr}        & &    & 756  & 42/45 &  & 0.78$^b$ & \textbf{0.85}$^\bullet$ &	0.84$^{m\circ}$  \\
\cmidrule{3-9}
\texttt{rosmap}      & & \multirow{2}{*}{\parbox{2cm}{\centering multiomics (numeric)}}  & 351  & 600 & \multirow{2}{*}{\parbox{2cm}{\centering euclidean, cosine}}  & 0.78 & \textbf{0.79}$^\circ$ & 0.76$^{m}$  \\
\texttt{brca}        & &  & 875  & 2503 &  & 0.89 & 0.91$^\circ$ & \textbf{0.91}$^{m\circ}$ \\
\cmidrule{3-9}
\texttt{wgs\_ovarian}  & & \multirow{4}{*}{\parbox{2cm}{\centering distributions, numeric}} & 219  & 56 & \multirow{4}{*}{\parbox{2cm}{\centering euclidean}} & 0.71$^{s}$ &	\textbf{0.76}$^\bullet$ &	0.71$^{s}$ \\
\texttt{wgs\_her2+tnbc}     & &  & 286  & 56 &  & 0.99$^{s\circ}$  & \textbf{0.99}$^\circ$  & 0.82$^{s}$ \\
\texttt{wgs\_her2+her2-}    & &  & 742 & 56 &  & \textbf{0.99}$^{s\circ}$ & 0.99$^\circ$ & 0.83$^{s}$ \\
\texttt{wgs\_her2-tnbc}    & &  & 780  & 56 &  & \textbf{0.90}$^{s\circ}$ & 0.90$^{\circ}$ & 0.76$^{s}$ \\
\bottomrule
\end{tabularx}
\begin{tablenotes}
\item [$\bullet$] result significantly better than both competitors
\item [$\circ$] result significantly better than one of the competitors
\item [$b$] RF used a bag of words representation of the data
\item [$m$] maximum performance obtained by SF out of all available distance measures
\item [$s$] RF and SF used data with distribution features encoded as sums
\end{tablenotes}
\label{tab:datasets}
\end{threeparttable}
\end{table}

The results clearly demonstrate that all three approaches are capable of producing high quality results on scalar data.
Nevertheless, there are some differences between the reported performance of the approaches.
For most datasets the results are very close, and either of the approaches (RF, SF, RSF) could be used.
However, on datasets \texttt{splice} and \texttt{svmguide3} RSF produces very similar results to RF, but both approaches significantly outperform SF by a large margin.
This outcome is probably due to the fact that although both RSF and SF are distance-based, the former still treats each feature separately, while the latter calculates distances over the whole feature space which, in this case, clearly obfuscates the decision boundary. The significantly worse performance of SF on the \texttt{splice} and \texttt{svmguide3} datasets showcases that it should be used with more caution than RF and RSF, which confirms recent findings of Czekalski and Morzy~\cite{Czekalski:2021}.

\subsection{Evaluation on Complex Data}
The second experiment was designed to assess the performance of Random Similarity Forests on complex data. This type of data should be a natural environment for Similarity Forests, so analogously to the first experiment, the aim of this comparison is to verify if RSF can match SF on this kind of data. Importantly, this is also the first reported test of Similarity Forests on actual complex data without explicitly defined features, since the algorithm was only tested on scalar data in the original paper~\cite{Sathe:2017} and the follow-up study~\cite{Czekalski:2021}.

In this group of experiments, both SF and RSF rely on the same set of distance measures for projections. For sequences of sets the algorithms use edit distance (with Jaccard distance as the cost function for element relabeling), for time series they use dynamic time warping, whereas for graphs the Ipsen-Mikhailov distance is used. We also wanted to include RF in this comparison, however, RF is incapable of processing varied sized sequences and graphs in their original form.
Therefore, for this classifier, we transform sequences of sets and graphs to a bag-of-words representation, where each sequence of sets/graph is represented as counts of items/nodes. Since time series were of the same length for a given dataset, RF could use individual time points as features, without any preprocessing. The results of this set of experiments are presented in Table~\ref{tab:datasets}.

We can observe that for the \texttt{items} dataset a simple bag-of-words representation is able to easily distinguish between the two classes, with RF significantly outperforming SF and RSF. This result clearly demonstrates that for easy problems, relying on approaches based on complex distance measures may actually be detrimental to the quality of predictions. In this case, the sequential information only obfuscates the true source of class information.
On the \texttt{length} dataset, where examples in the two classes are differentiated only by the lengths of the sequences, RSF and SF both significantly outperformed RF. This is expected as the information about the length of the sequences is very easy to capture through edit distance.
Finally, on the \texttt{order} dataset, where the elements in one of the classes are ordered while in the other they are not, RF was barely able to find any regularities and was significantly outperformed by both SF and RSF.

Results for real-world time series and graphs confirm the above observations. On each time series and graph dataset RSF and SF significantly outperformed RF. The differences between RSF and SF were much smaller and, with the exception of the \texttt{proteins} dataset, were not statistically significant.

\subsection{Evaluation on Mixed Data}
The final set of experiments was designed to evaluate Random Similarity Forests on datasets with a mix of simple numerical features and complex object-like features. These are the types of datasets that neither Random Forest nor Similarity Forest can process out of the box. Here, RF used only the numerical features, whereas SF treated the examples as complex objects. In cases where multiple distance measures were available for the data (time series, graphs, multiomics), RSF used each measure as an independent feature whereas SF used each of them individually. Therefore, for each dataset with multiple distances, SF was ran multiple times, each time with a different measure. The best performing distance was later taken into account during statistical tests.

The results clearly show that RSF is the best choice for mixed data. For 7 out of 10 datasets RSF achieved the best performance and was never statistically significantly worse than any other approach. It is worth noting that on the \texttt{bzr} and \texttt{wgs\_ovarian} datasets, RSF outperforms RF and SF by a large margin. These are two stark examples that on some datasets the combination of numeric and complex data can offer much more than any of the component representations alone, and that multiomics data are potentially a good training ground for algorithms like RSF. Interestingly, there is no clear runner-up in this set of experiments. On the \texttt{beetlefly}, \texttt{birdchicken}, \texttt{rosmap}, and \texttt{wgs\_} datasets SF is the significantly underperforming algorithm, whereas on \texttt{bzr},  \texttt{dhfr}, \texttt{brca} RF takes last spot. This suggests that depending on the application domain, numeric or complex features play a more important role. Algorithms like RSF can take advantage of this property and steer towards the more rewarding feature types for a given dataset.

\subsection{Summary of the Results}
The main takeaways from the experiments are as follows.
\begin{itemize}
    \item Due to the increased randomness, Random Similarity Forests usually need to evaluate more features (\textit{max\_features}) during splits than Random Forests.
    \item The proposed algorithm was able to match the performance of Random Forests and match or outperform Similarity Forests on datasets consisting of scalar features.
    \item Random Similarity Forest was also able to match or outperform Similarity Forests and outperform Random Forests on complex data.
    \item Random Similarity Forest was the only algorithm capable of using a mixture of scalar and complex features. This property helped Random Similarity Forests outperform Random Forests and Similarity Forests on datasets with both types of features. The experiments have shown that for real-world omics data transforming complex features to scalars or using a single example-wide distance measure may result in information loss.
\end{itemize}

\section{Conclusions}
\label{sec:conclusions}
In this paper, we have addressed the problem of classifying data with a mixture of scalar and complex features by proposing a new classifier, called Random Similarity Forest.
Like other decision forest algorithms, such as Random Forests and Similarity Forests, the proposed algorithm relies on bagging of decision trees, but in contrast to other approaches splits the tree nodes according to distance-based projections of single feature values.
This allows for a very flexible classification approach, in which traditional numerical features can be used alongside complex object-like features with domain-specific distance measures, taking full advantage of each data type.
The proposed method was experimentally evaluated against Random Forests and Similarity Forests on datasets with scalar, complex, and mixed data features, showcasing its capabilities across all possible data domain configurations.

Experimental results on 30 datasets show that Random Similarity Forests can be considered a safe alternative to the well-established Random Forests on scalar datasets and are on par or exceed the performance of Similarity Forests on complex data. Most importantly, however, the experiments show that the proposed classifier was the only one inherently capable of dealing with datasets consisting of a mixture of scalar and complex features, and outperformed both of the alternatives on most datasets with mixed feature types. In situations were numeric and distance features offer complementary information, Random Similarity Forests can achieve results that are not attainable by the individual feature types alone. In situations were only numeric or only complex features are informative, the proposed algorithm achieves results that are close to the algorithm using only the informative representation.

The method proposed in this paper opens many new avenues for further research. Regarding the classifier itself, it would be interesting to verify the impact of different proportions of numeric and complex features on classifier performance. To control this effect, one could implement a weighting mechanism that influences how many features of each type are taken into account at a given tree node split. Moreover, the effects of other forms of randomization on the proposed algorithm, such as random splits in the projection space, could also be inspected. Finally, it will be interesting to observe what types of features and distance measures are most useful when  applied to different types of multiomics data, which seem to be one of the most promisisng application domains for Random Similarity Forests.

\bibliographystyle{splncs04}
\bibliography{rsf}

\end{document}